\documentclass{article}
\usepackage[T1]{fontenc}

\usepackage[utf8]{inputenc} 
\usepackage[T1]{fontenc}
\usepackage{tgtermes}
\usepackage{amssymb}
\usepackage{amsfonts}

\usepackage{tgtermes}
\usepackage{amsmath}
\usepackage{amsthm}  
\usepackage{scalefnt,letltxmacro}
\LetLtxMacro{\oldtextsc}{\textsc}
\renewcommand{\textsc}[1]{\oldtextsc{\scalefont{1.10}#1}}
\usepackage[scaled=0.92]{PTSans}

\usepackage[usenames,dvipsnames]{xcolor}
\definecolor{shadecolor}{gray}{0.9}

\usepackage[parfill]{parskip}
\usepackage{afterpage}
\usepackage{framed}
\usepackage{nicefrac}
\usepackage{bm}
\usepackage{fixmath}

\usepackage[colorinlistoftodos,
           prependcaption,
           textsize=small,
           backgroundcolor=yellow,
           linecolor=lightgray,
           bordercolor=lightgray]{todonotes}

\usepackage{lineno}

\usepackage{ragged2e}

\DeclareRobustCommand{\parhead}[1]{\textbf{#1}~}

\usepackage{graphicx}
\usepackage[labelfont=it,labelsep=period]{caption}

\usepackage{booktabs}
\usepackage{arydshln} 
\usepackage{multirow}

\usepackage{natbib}

\usepackage{listings}
\usepackage{fancyvrb}
\fvset{fontsize=\normalsize}

\usepackage[colorlinks,linktoc=all]{hyperref}
\usepackage[all]{hypcap}
\hypersetup{citecolor=Violet}
\hypersetup{linkcolor=black}
\hypersetup{urlcolor=MidnightBlue}
\usepackage{url}

\usepackage[nameinlink]{cleveref}
\creflabelformat{equation}{#1#2#3}
\crefname{equation}{eq.}{eqs.}  
\Crefname{equation}{Eq.}{Eqs.}

\usepackage[acronym,smallcaps,nowarn]{glossaries}

\usepackage{listings}
\lstdefinestyle{alp_style}{
    commentstyle=\color{OliveGreen},
    numberstyle=\tiny\color{black!60},
    stringstyle=\color{BrickRed},
    basicstyle=\ttfamily\scriptsize,
    breakatwhitespace=false,
    breaklines=true,
    captionpos=b,
    keepspaces=true,
    numbers=none,
    numbersep=5pt,
    showspaces=false,
    showstringspaces=false,
    showtabs=false,
    tabsize=2
}
\usepackage{lipsum}

\DeclareRobustCommand{\E}[2]{\mathbb{E}_{#1}\left[#2\right]}

\newcommand{\g}{\, | \,}
\newcommand{\prm}{\, ; \,}

\newcommand{\Lcal}{\mathcal{L}}
\newcommand{\Ncal}{\mathcal{N}}

\newcommand{\Naturals}{\mathbb{N}}

\newcommand{\Reals}{\mathbb{R}}

\newacronym{ADVI}{advi}{automatic differentiation variational inference}
\newacronym{AR}{a{\small\&}r}{augment and reduce}

\newacronym{BBVI}{bbvi}{black-box variational inference}

\newacronym{CDF}{cdf}{cumulative distribution function}
\newacronym{CS-EFE}{cs-efe}{context selection for exponential family embeddings}
\newacronym{CTM}{ctm}{correlated topic model}

\newacronym[\glslongpluralkey={deep exponential families}]{DEF}{def}{deep exponential family}
\newacronym{DMIS}{dmis}{deterministic multiple importance sampling}

\newacronym{EFE}{efe}{exponential family embeddings}
\newacronym{ELBO}{elbo}{evidence lower bound}
\newacronym{EM}{em}{expectation maximization}
\newacronym{ETM}{etm}{embedded topic model}

\newacronym{GNTS}{gn-ts}{gamma-normal time series model}
\newacronym{G-REP}{g-rep}{generalized reparameterization}

\newacronym{HMC}{hmc}{{H}amiltonian {M}onte {C}arlo}

\newacronym{KL}{kl}{{K}ullback-{L}eibler}

\newacronym{LDA}{lda}{latent {D}irichlet allocation}

\newacronym{MAP}{map}{\emph{maximum a posteriori}}
\newacronym{MCMC}{mcmc}{{M}arkov chain {M}onte {C}arlo}
\newacronym{MF}{mf}{matrix factorization}
\newacronym{MIS}{mis}{multiple importance sampling}
\newacronym{ML}{ml}{maximum likelihood}

\newacronym{NVDM}{nvdm}{neural variational document model}

\newacronym{OBBVI}{o-bbvi}{overdispersed black-box variational inference}
\newacronym{OVE}{ove}{one-vs-each}

\newacronym{PPCA}{ppca}{probabilistic principal component analysis}

\newacronym{SIVI}{sivi}{semi-implicit variational inference}
\newacronym{SVI}{svi}{stochastic variational inference}
\newacronym{SMC}{smc}{sequential Monte Carlo}

\newacronym{TMES}{tmes}{topic model in embedding space}

\newacronym{USIVI}{uivi}{unbiased implicit variational inference}

\newacronym{VAE}{vae}{variational autoencoder}
\newacronym{VCD}{vcd}{variational contrastive divergence}
\newacronym{VEM}{vem}{variational expectation maximization}
\newacronym{VI}{vi}{variational inference}

\usepackage{microtype}
\usepackage{graphicx}
\usepackage{subfigure}
\usepackage{booktabs} 

\usepackage{hyperref}

\usepackage[accepted]{icml2019}

\hypersetup{citecolor=Violet}
\hypersetup{linkcolor=black}
\hypersetup{urlcolor=MidnightBlue}

\icmltitlerunning{A Contrastive Divergence for Combining Variational Inference and MCMC}

\begin{document}

\twocolumn[
\icmltitle{A Contrastive Divergence for Combining Variational Inference and MCMC}



\icmlsetsymbol{equal}{*}

\begin{icmlauthorlist}
	\icmlauthor{Francisco J.\ R.\ Ruiz}{ca,cu}
	\icmlauthor{Michalis K.\ Titsias}{dm}
\end{icmlauthorlist}

\icmlaffiliation{cu}{Columbia University, New York, USA}
\icmlaffiliation{ca}{University of Cambridge, Cambridge, UK}
\icmlaffiliation{dm}{DeepMind, London, UK}

\icmlcorrespondingauthor{Francisco J.\ R.\ Ruiz}{f.ruiz@columbia.edu}

\icmlkeywords{variational inference, Markov chain Monte Carlo, latent variable models}

\vskip 0.3in
]



\printAffiliationsAndNotice{}  

\begin{abstract}
We develop a method to combine \gls{MCMC} and \gls{VI}, leveraging the advantages of both inference approaches. Specifically, we improve the variational distribution by running a few \gls{MCMC} steps. To make inference tractable, we introduce the \gls{VCD}, a new divergence that replaces the standard \gls{KL} divergence used in \gls{VI}. The \gls{VCD} captures a notion of discrepancy between the initial variational distribution and its improved version (obtained after running the \gls{MCMC} steps), and it converges asymptotically to the symmetrized \gls{KL} divergence between the variational distribution and the posterior of interest. The \gls{VCD} objective can be optimized efficiently with respect to the variational parameters via stochastic optimization. We show experimentally that optimizing the \gls{VCD} leads to better predictive performance on two latent variable models: logistic matrix factorization and \glspl{VAE}.
\end{abstract}

\section{Introduction}
\label{sec:introduction}
\glsresetall

\Gls{VI} and \gls{MCMC} are two of the main approximate Bayesian inference methods \citep{Bishop2006,Murphy2012}. While \gls{MCMC} is asymptotically exact, \gls{VI} enjoys other advantages: \gls{VI} is typically faster, makes it easier to assess convergence, and enables amortized inference---a way to quickly approximate the posterior over the local latent variables.

A natural question is whether it is possible to combine \gls{MCMC} and \gls{VI} to leverage the advantages of each inference method. Such topic has attracted a lot of attention in the recent literature \citep[see, e.g.,][]{Salimans2015,Maddison2017,Naesseth2018,Le2018,Hoffman2017,Li2017approximate}.

We develop a method for combining \gls{VI} and \gls{MCMC} that improves an explicit
variational distribution (i.e., with analytic density) by applying \gls{MCMC} sampling.
The method runs a few iterations of an \gls{MCMC} chain initialized with a sample from the explicit distribution, so that each \gls{MCMC} step successively improves the initial distribution.

Combining \gls{VI} and \gls{MCMC} in this way is challenging. Specifically, fitting the parameters of the explicit variational distribution is intractable under the standard \gls{VI} framework.
This is because the improved distribution, obtained by \gls{MCMC} sampling, is defined implicitly, i.e., its density cannot be evaluated.
Thus, we cannot minimize the \gls{KL} divergence between the improved variational distribution and the true posterior of interest.

To address this challenge, we develop a divergence that allows combining \gls{VI} and \gls{MCMC} in a principled manner. We refer to it as \emph{\acrlong{VCD}} (\acrshort{VCD}).
The \acrshort{VCD} replaces the standard \gls{KL} objective of \gls{VI}, enabling tractable optimization. The key property of the \acrshort{VCD} is that it is possible to obtain unbiased estimates of its gradient to perform stochastic optimization \citep{Robbins1951}. The \acrshort{VCD} differs from the standard \gls{KL} in that it captures a notion of discrepancy between the improved and the initial variational distributions. The properties of the \acrshort{VCD} make it a valid objective function for Bayesian inference: it is non-negative and it becomes zero only when the variational distribution matches the posterior.

Additionally, as the number of \gls{MCMC} steps increases, the \acrshort{VCD} converges asymptotically to the symmetrized \gls{KL} between the initial explicit variational distribution and the posterior. This implies that the variational distributions fitted by minimizing the \acrshort{VCD} will exhibit larger variance than the distributions fitted with the standard \gls{KL} divergence, even for moderate values of the number of \gls{MCMC} iterations.

We fit the variational parameters of the initial distribution by following stochastic gradients of the \acrshort{VCD}.
In contrast to the method of \citet{Hoffman2017} (which optimizes a different objective), the stochastic gradients of the \acrshort{VCD} depend on the improved \gls{MCMC} samples; therefore, the \gls{MCMC} samples provide feedback to the optimization of the variational parameters.
Unlike the method of \citet{Li2017approximate}, the \acrshort{VCD} leads to stable optimization regardless of the number of \gls{MCMC} steps, since it is a well-defined divergence.

We demonstrate the \acrshort{VCD} on latent variable models, where there is a local latent variable corresponding to each observation, and the goal is to fit some global model parameters via approximate \gls{ML}. In this setting, amortized inference allows us to obtain a quick approximation of the posterior over each latent variable. We refine the amortized variational distribution using \gls{MCMC}, and we use the \acrshort{VCD} objective to fit the parameters of the amortized distribution. We show experimentally that the models fitted with the \acrshort{VCD} objective have better predictive performance than when we use alternative approaches, including standard \gls{VI} and the method of \citet{Hoffman2017}.


\section{Method Description}
\label{sec:method}
\glsresetall

Here we formally describe the method for refining the variational distribution with \gls{MCMC} sampling, as well as the \gls{VCD} that enables stochastic optimization.

\Cref{subsec:background} provides a brief background on \gls{VI} and introduces the notation. \Cref{subsec:refining} describes the improved distribution that uses \gls{MCMC} sampling. \Cref{subsec:divergence} introduces the \gls{VCD}, and \Cref{subsec:gradients} describes how to form unbiased estimators of its gradient. Finally, \Cref{subsec:full_algorithm} summarizes the full algorithm.

\subsection{Background: Variational inference}
\label{subsec:background}

Consider the joint distribution $p(x,z)$ over the data $x$ and latent variables $z$. We are interested in approximating the posterior $p(z\g x)$. Consider a variational approximation $q_{\theta}(z)$, a family of distributions with parameter $\theta$. The distribution $q_{\theta}(z)$ may also depend on $x$ as in amortized \gls{VI}, where $q_{\theta}(z)=q_{\theta}(z\g x)$. To avoid clutter, we simply write $q_{\theta}(z)$ throughout this section.

The variational parameter $\theta$ is fitted by minimizing the \gls{KL} divergence between the variational distribution $q_{\theta}(z)$ and the posterior $p(z\g x)$, $\textrm{KL}(q_{\theta}(z) \;||\; p(z\g x))$. This is equivalent to maximizing the \gls{ELBO},
\begin{equation}\label{eq:elbo}
	\Lcal_{\textrm{standard}}(\theta) = \E{q_{\theta}(z)}{f_{\theta}(z)},
\end{equation}
where we use the shorthand notation $f_{\theta}(z)$ for the argument of the expectation, which we call the ``instantaneous \gls{ELBO},''
\begin{equation}\label{eq:instantaneous_elbo}
	f_{\theta}(z) \triangleq \log p(x,z) - \log q_{\theta}(z).
\end{equation}

\subsection{Refining the variational approximation}
\label{subsec:refining}

We improve the variational distribution $q_{\theta}(z)$ by running an \gls{MCMC} method initialized at $q_{\theta}(z)$ and whose stationary distribution is the posterior $p(z\g x)$. Suppose that we apply such a Markov chain for a fixed number of iterations $t\in\Naturals$. This results in a marginal distribution over the latent variables that we denote $q_{\theta}^{(t)}(z)$,
\begin{equation}\label{eq:def_improved_q}
	q_{\theta}^{(t)}(z) = \int Q^{(t)}(z\g z_0) q_{\theta}(z_0) dz_0,
\end{equation}
where $Q^{(t)}(z\g z_0)$ denotes the overall transition kernel that takes an initial sample from $q_{\theta}(z_0)$ and after $t$ iterations it produces a sample from $q_{\theta}^{(t)}(z)$.

The distribution $q_{\theta}^{(t)}(z)$ is an \emph{improvement} of the variational distribution $q_{\theta}(z)$ for any $t$ because it is closer to the posterior $p(z\g x)$ in terms of \gls{KL} divergence \citep[][p.\ 81]{Cover2006}, i.e.,
\begin{equation}\label{eq:improvement_definition1}
	\textrm{KL}(q_{\theta}(z) \;||\; p(z\g x)) \geq \textrm{KL}(q_{\theta}^{(t)}(z) \;||\; p(z\g x)).
\end{equation}
If $q_{\theta}(z) \neq p(z\g x)$, then $q_{\theta}^{(t)}(z)$ is strictly closer to the posterior 
and the above becomes a strict inequality.  In the case where $q_{\theta}(z) = p(z\g x)$, then the \gls{KL} divergence is zero and cannot be reduced, therefore the \gls{KL} divergence between $q_{\theta}^{(t)}(z)$ and the posterior is also zero,
\begin{equation}\label{eq:improvement_definition2}
	\textrm{KL}(q_{\theta}(z) \;||\; p(z\g x)) = \textrm{KL}(q_{\theta}^{(t)}(z) \;||\; p(z\g x)) = 0.
\end{equation}
We now develop a triangle inequality that will play an important role in the development of the divergence in \Cref{subsec:divergence}. Specifically, we add the non-negative term $\textrm{KL}(q_{\theta}^{(t)}(z) \;||\; q_{\theta}(z))$ to the left-hand side of \Cref{eq:improvement_definition1},
\begin{equation}\label{eq:triangle_inequality}
	\begin{split}
		& \textrm{KL}(q_{\theta}(z) \;||\; p(z\g x)) + \textrm{KL}(q_{\theta}^{(t)}(z) \;||\; q_{\theta}(z)) \\ 
		& \geq \textrm{KL}(q_{\theta}^{(t)}(z) \;||\; p(z\g x)).
	\end{split}
\end{equation}
Triangle inequalities do not hold in general for \gls{KL} divergences (since the \gls{KL} is not a norm), but \Cref{eq:triangle_inequality} holds because $q_{\theta}^{(t)}(z)$ is an improvement of $q_{\theta}(z)$ with the respect to the target $p(z\g x)$. 

While the original variational distribution $q_{\theta}(z)$ is typically tractable and allows for direct maximization of the \gls{ELBO} in \Cref{eq:elbo}, we cannot directly use the improved distribution $q_{\theta}^{(t)}(z)$ as the variational approximation. The reason is that the resulting \gls{ELBO}, given by
\begin{equation}\label{eq:elbo_improved}
	\Lcal_{\textrm{improved}}(\theta) = \E{q_{\theta}^{(t)}(z)}{\log p(x,z) - \log q_{\theta}^{(t)}(z)},
\end{equation}
presents several challenges. First, the log-density of the improved distribution $\log q_{\theta}^{(t)}(z)$ cannot be evaluated because it is implicitly defined through \Cref{eq:def_improved_q}. This
makes standard stochastic optimization techniques for Monte Carlo-based maximization of \Cref{eq:elbo_improved}  intractable to apply. 
Second, the optimization of \Cref{eq:elbo_improved} can be difficult because $q_{\theta}^{(t)}(z)$ depends more weakly on the variational parameters $\theta$ than the initial $q_{\theta}(z)$, since the former is closer to the posterior $p(z\g x)$, which is independent of $\theta$. In fact, in the limit when the number of \gls{MCMC} steps is very large ($t\rightarrow\infty$), the objective $\Lcal_{\textrm{improved}}(\theta)$ becomes independent of $\theta$. Thus, it is challenging to use the improved distribution as the variational distribution directly.

We next introduce a divergence that can be tractably optimized while avoiding these two challenges.

\subsection{The variational contrastive divergence}
\label{subsec:divergence}

Our goal is to find an alternative divergence that can be tractably and efficiently optimized over $\theta$. The divergence we wish to find needs to satisfy two conditions: (i) it must be non-negative for any value of the variational parameters $\theta$, and (ii) it must become zero only when the variational distribution $q_{\theta}(z)$ matches the posterior $p(z\g x)$.

A first idea to form such a divergence is to start from the inequality in \Cref{eq:improvement_definition1}, bring 
everything to the left-hand side, and express the discrepancy between $q_{\theta}(z)$ and its improvement $q_{\theta}^{(t)}(z)$ in terms of the difference
\begin{equation}\label{eq:l_diff}
	\Lcal_{\textrm{diff}}(\theta) = \textrm{KL}(q_{\theta}(z) \;||\; p(z\g x)) - \textrm{KL}(q_{\theta}^{(t)}(z) \;||\; p(z\g x)).
\end{equation}
This is a proper divergence since it satisfies the two criteria outlined above. It takes non-negative values because $q_{\theta}^{(t)}(z)$ reduces the \gls{KL} divergence, and it becomes zero only when $q_{\theta}(z)=p(z\g x)$, as discussed in \Cref{subsec:refining}.

However, the objective $\Lcal_{\textrm{diff}}(\theta)$ is still intractable to minimize due to the term $\log q_{\theta}^{(t)}(z)$ that appears in a similar way as in \Cref{eq:elbo_improved}.
To address that, we make use of the triangle inequality 
in \Cref{eq:triangle_inequality} and modify the divergence above by adding the regularization term $\textrm{KL}(q_{\theta}^{(t)}(z) \;||\; q_{\theta}(z))$, which adds an extra force for reducing the discrepancy between the initial distribution and its improvement. This leads to the \gls{VCD} divergence,
\begin{equation}\label{eq:new_divergence}
	\Lcal_{\textrm{VCD}}(\theta) \triangleq \Lcal_{\textrm{diff}}(\theta) + \textrm{KL}(q_{\theta}^{(t)}(z) \;||\; q_{\theta}(z)).
\end{equation}
The \gls{VCD} is also a proper divergence. It is non-negative due to the triangle inequality (see \Cref{eq:triangle_inequality}) and it becomes zero only when $q_{\theta}(z)$ matches the posterior $p(z\g x)$.

The \gls{VCD} in \Cref{eq:new_divergence} is tractable, in the sense that we can obtain unbiased stochastic gradients with respect to $\theta$ without evaluating the density of the improved distribution $q_{\theta}^{(t)}(z)$. In fact, unlike most common divergences,  we
can also obtain unbiased stochastic estimates
of the actual value of the \gls{VCD}.    
The reason is that the problematic term $\log q_{\theta}^{(t)}(z)$ now cancels out. To see that, we rearrange the terms in \Cref{eq:new_divergence} (see Appendix~1) and express the divergence as
\begin{equation}\label{eq:new_divergence_rearranged}
	\Lcal_{\textrm{VCD}}(\theta) = -\E{q_{\theta}(z)}{f_{\theta}(z)} + \E{q_{\theta}^{(t)}(z)}{f_{\theta}(z)}.
\end{equation}
The \gls{VCD} contains two terms. The first term is the negative standard \gls{ELBO} (\Cref{eq:elbo}), which involves only the tractable distribution $q_{\theta}(z)$. The second term is also an expectation of the instantaneous \gls{ELBO} (\Cref{eq:instantaneous_elbo}), taken with respect to the improved distribution $q_{\theta}^{(t)}(z)$ instead. Even though the expectation is taken with respect to $q_{\theta}^{(t)}(z)$, the argument of the expectation no longer involves the improved distribution. This allows us to form Monte Carlo estimates of the gradient of $\Lcal_{\textrm{VCD}}(\theta)$ with respect to the variational parameters, as detailed in \Cref{subsec:gradients}.

\parhead{Asymptotic properties of the \acrshort{VCD}.}
In the limit when $t\rightarrow\infty$, the improved distribution $q_{\theta}^{(t)}(z)$ converges to the posterior $p(z\g x)$, and the divergence $\Lcal_{\textrm{VCD}}(\theta)$ converges to the symmetrized \gls{KL} divergence between the variational distribution and the posterior,\footnote{This can be seen by substituting $q_{\theta}^{(t)}(z)=p(z\g x)$ in \Cref{eq:new_divergence}.}
$\textrm{KL}_{\textrm{sym}}(q_{\theta}(z) \;||\; p(z\g x)) = \textrm{KL}(q_{\theta}(z) \;||\; p(z\g x)) + \textrm{KL}(p(z\g x) \;||\; q_{\theta}(z))$. This ensures that $\Lcal_{\textrm{VCD}}(\theta)$ depends on the variational parameters even when the number of \gls{MCMC} steps is large, unlike the objective in \Cref{eq:elbo_improved}. Moreover, the \gls{VCD} favors variational distributions $q_{\theta}(z)$ with larger variance than the standard \gls{ELBO}, as it converges to the symmetrized \gls{KL} divergence.

We can form a generalization of the \gls{VCD} that interpolates between the standard and the symmetrized \gls{KL} according to a parameter $\alpha$ (see Appendix~2). The experimentation of this generalization is left for future work.


\subsection{Taking gradients of the \acrshort{VCD}}
\label{subsec:gradients}

We now show how to estimate the gradient of the \gls{VCD} with respect to $\theta$.
The first term in \Cref{eq:new_divergence_rearranged} is the negative standard \gls{ELBO}, for which we can obtain unbiased gradients with respect to $\theta$ by using either the score function estimator (or \textsc{reinforce}) \citep{Carbonetto2009,Paisley2012,Ranganath2014} or the reparameterization gradient \citep{Rezende2014,Titsias2014_doubly,Kingma2014}. Assuming that $q_{\theta}(z)$ is reparameterizable as $\varepsilon\sim q(\varepsilon)$, $z=h_{\theta}(\varepsilon)$, then the reparameterization gradient of the (negative) first term is
\begin{equation}\label{eq:gradient_term1}
	\nabla_{\theta} \E{q_{\theta}(z)}{f_{\theta}(z)} = \E{q(\varepsilon)}{ \nabla_z f_{\theta}(z) \big|_{z=h_{\theta}(\varepsilon)} \times \nabla_{\theta} h_{\theta}(\varepsilon)},
\end{equation}
which can be estimated with samples $\varepsilon\sim q(\varepsilon)$.

We now focus on the non-standard second term. We express its gradient as an expectation with respect to the improved distribution $q_{\theta}^{(t)}(z)$, from which we can sample, thus enabling stochastic optimization. To achieve that, we write the gradient of the second term in \Cref{eq:new_divergence_rearranged} as the sum of two expectations with respect to $q_{\theta}^{(t)}(z)$,
\begin{equation}\label{eq:gradient_term2}
	\begin{split}
		& \nabla_{\theta} \E{q_{\theta}^{(t)}(z)}{f_{\theta}(z)} = - \E{q_{\theta}^{(t)}(z)}{\nabla_{\theta} \log q_{\theta}(z)} \\
		& + \mathbb{E}_{q_{\theta}(z_0)}\left[\mathbb{E}_{Q^{(t)}(z\g z_0)}\!\left[f_{\theta}(z)\right]\nabla_{\theta}\!\log q_{\theta}(z_0)\right].
	\end{split}
\end{equation}
This expression allows us to form Monte Carlo estimates of the gradient using samples from $q_{\theta}^{(t)}(z)$ and does not require to evaluate the intractable log-density $\log q_{\theta}^{(t)}(z)$.
Samples from $q_{\theta}^{(t)}(z)$ can be obtained by first sampling $z_0\sim q_{\theta}(z)$ and then running $t$ \gls{MCMC} steps, $z\sim Q^{(t)}(z\g z_0)$.

\parhead{Proof of \Cref{eq:gradient_term2}.}
We now show how to derive \Cref{eq:gradient_term2}. We first apply the product rule for derivatives,
\begin{equation}\label{eq:proof_line1}
	\begin{split}
		& \nabla_{\theta} \E{q_{\theta}^{(t)}(z)}{f_{\theta}(z)} = \int q_{\theta}^{(t)}(z) \times \nabla_{\theta} f_{\theta}(z) dz \\
		&+ \int \nabla_{\theta} q_{\theta}^{(t)}(z) \times f_{\theta}(z) dz.
	\end{split}
\end{equation}
The first integral is straightforward to unbiasedly approximate by drawing samples from $q_{\theta}^{(t)}(z)$, and hence it is directly one of the terms in \Cref{eq:gradient_term2}. Note that, since the model $p(x,z)$ does not depend on $\theta$, the gradient of the instantaneous \gls{ELBO} is $\nabla_{\theta} f_{\theta}(z)=-\nabla_{\theta}\log q_{\theta}(z)$.

For the term $\nabla_{\theta} q_{\theta}^{(t)}(z)$ in the second integral, we substitute the definition of the improved distribution in \Cref{eq:def_improved_q} and apply the log-derivative trick, yielding
\begin{equation}\label{eq:proof_line2}
	\begin{split}
		& \nabla_{\theta} q_{\theta}^{(t)}(z) = \nabla_{\theta} \int Q^{(t)}(z\g z_0) q_{\theta}(z_0) dz_0 \\
		& = \int Q^{(t)}(z\g z_0) q_{\theta}(z_0) \nabla_{\theta} \log q_{\theta}(z_0) dz_0.
	\end{split}
\end{equation}
Here we have made use of the fact that the $t$-step \gls{MCMC} kernel $Q^{(t)}(z\g z_0)$ is independent of $\theta$. Finally, we obtain \Cref{eq:gradient_term2} by substituting \Cref{eq:proof_line2} into \Cref{eq:proof_line1}. 
\hfill\rlap{\hspace*{2em}$\square$}

\parhead{Controlling the variance.}
Since the estimator based on (the last line of) \Cref{eq:gradient_term2} is a score function estimator, it may suffer from high variance. We form a simple control variate to reduce the variance, obtained as an exponentially decaying average of the previous stochastic values.

More in detail, consider the term in the last line of \Cref{eq:gradient_term2}, $\E{ q_{\theta}(z_0)}{ w_{\theta}(z_0) \times \nabla_{\theta} \log q_{\theta}(z_0)}$, where we have defined $w_{\theta}(z_0)\triangleq \E{Q^{(t)}(z\g z_0)}{f_{\theta}(z)}$. This expectation can be equivalently written using a control variate $C$ as $\E{ q_{\theta}(z_0)}{ \left( w_{\theta}(z_0) - C \right) \times \nabla_{\theta} \log q_{\theta}(z_0)}$, as long as $C$ does not depend on $z_0$. We set $C$ as an exponentially decaying average of the \emph{previous} stochastic values of $w_{\theta}(z_0)$, i.e., the values from previous iterations of gradient descent. At a given iteration of gradient descent, the one-sample stochastic estimate for $w_{\theta}(z_0)$ is simply $f_{\theta}(z)$.

Finally, note that when the number of \gls{MCMC} iterations $t$ is very large, the resulting estimator based on \Cref{eq:gradient_term2} has much lower variance, and control variates might not be needed. The reason is that the distribution $q_{\theta}^{(t)}(z)$ becomes less dependent on $\theta$, and is gradient (\Cref{eq:proof_line2}) becomes zero. In that case, \Cref{eq:gradient_term2} simplifies as $\nabla_{\theta} \E{q_{\theta}^{(t)}(z)}{f_{\theta}(z)} \approx - \E{q_{\theta}^{(t)}(z)}{\nabla_{\theta} \log q_{\theta}(z)}$, and the resulting estimator based on this term only has much lower variance than the estimator based on the two terms.

\subsection{Full algorithm}
\label{subsec:full_algorithm}

We now summarize the full algorithm to minimize the \gls{VCD} from \Cref{subsec:divergence}. The algorithm forms a one-sample estimator of the gradient of the \gls{VCD}. For that, it first samples $z_0\sim q_{\theta}(z_0)$ from the initial distribution (this can be done using reparameterization) and then runs $t$ \gls{MCMC} steps to obtain the improved sample $z\sim Q^{(t)}(z\g z_0)$.

The algorithm uses the initial sample $z_0$ to form an unbiased estimator of the gradient of the standard term, $\nabla_{\theta} \E{q_{\theta}(z_0)}{f_{\theta}(z_0)}$, using \Cref{eq:gradient_term1} (alternatively, the score function estimator can be used instead). With both samples $z_0$ and $z$, the algorithm makes use of \Cref{eq:gradient_term2} to estimate the non-standard gradient $\nabla_{\theta} \E{q_{\theta}^{(t)}(z)}{f_{\theta}(z)}$. The control variate $C$ is used in this step to reduce the variance of the estimator. Finally, the two terms are combined following \Cref{eq:new_divergence_rearranged} to obtain the gradient estimator of the divergence.

The full procedure is given in \Cref{alg:full_algorithm}.\footnote{Code is available online at \url{https://github.com/franrruiz/vcd_divergence}.}
 It has two additional parameters.
The first one is the decay parameter $\gamma$ for the updates of the control variate. We set $\gamma=0.9$. The second parameter is the stepsize $\rho$. We set the stepsize using RMSProp \citep{Tieleman2012}; at each iteration $\ell$ we set $\rho^{(\ell)}=\eta /(1+\sqrt{G^{(\ell)}})$, where $\eta$ is the learning rate, and the updates of $G^{(\ell)}$ depend on the gradient estimate $\widehat{\nabla}_{\theta} \Lcal_{\textrm{VCD}}^{(\ell)}$ as $G^{(\ell)}=0.9G^{(\ell-1)} + 0.1 (\widehat{\nabla}_{\theta} \Lcal_{\textrm{VCD}}^{(\ell)})^2$.


\section{Related Work}
\label{sec:related}
\glsresetall

There are several related works in the literature. For example, \citet{Salimans2015} combine \gls{MCMC} and \gls{VI} by introducing auxiliary variables (associated with the \gls{MCMC} iterations) that need to be inferred together with the rest of the variables. Other works use rejection sampling within the variational framework \citep{Naesseth2017,Grover2018} or meld sequential Monte Carlo and \gls{VI} \citep{Maddison2017,Naesseth2018,Le2018}.

More related to ours is the work of \citet{Hoffman2017, Li2017approximate, Zhang2018, Titsias2017, Habib2019}.
Specifically, the method of \citet{Hoffman2017} performs 
approximate \gls{ML} estimation in non-linear latent variable models based on the \gls{VAE}.
The \textsc{e}-step of the approximate \gls{ML} procedure minimizes 
the standard \gls{KL} divergence  $\textrm{KL}(q_{\theta}(z) \;||\; p(z\g x))$, where $q_{\theta}(z)$ is an explicit amortized distribution;
and the \textsc{m}-step updates the model parameters using an improved \gls{MCMC} distribution $q_{\theta}^{(t)}(z)$. 
Since it uses the standard \gls{KL} divergence of \gls{VI}, the \gls{MCMC} procedure does not provide feedback 
when learning the variational parameters $\theta$.

The amortized \gls{MCMC} technique of \citet{Li2017approximate}
incorporates feedback from \gls{MCMC} back to the 
parameters of the explicit distribution. Their method
aims at learning $\theta$ by minimizing the \gls{KL} divergence between the improved 
distribution and its initialization, i.e.,
$\textrm{KL}(q_{\theta}^{(t)}(z) \;||\;  q_{\theta}(z) )$. This divergence
is intractable, and \citet{Li2017approximate} approximate its gradient 
$\nabla_{\theta} \textrm{KL}(q_{\theta}^{(t)}(z) \;||\;  q_{\theta}(z) )\approx \E{q_{\theta}^{(t)}(z)}{\nabla_{\theta} \log q_{\theta}(z)}$, which ignores the 
dependence of $q_{\theta}^{(t)}(z)$ on $\theta$. Subsequently, this can lead to   
unstable optimization over $\theta$, especially for small values of $t$, 
when $q_{\theta}^{(t)}(z)$ strongly depends on $\theta$. The procedure becomes stable for large $t$, when the \gls{MCMC} chain converges and 
$q_{\theta}^{(t)}(z) = p(z\g x)$.
Notice that the expectation $ \E{q_{\theta}^{(t)}(z)}{\nabla_{\theta} \log q_{\theta}(z)}$
appears also in our approach (see \Cref{eq:gradient_term2}), but it is just one part of the overall gradient for optimizing $\Lcal_{\textrm{VCD}}(\theta)$,
a divergence that leads to stable optimization.


\begin{algorithm}[tb]
	\caption{Minimization of the \acrshort{VCD}}
	\label{alg:full_algorithm}
	\begin{algorithmic}
		\STATE {\bfseries Input:} data $x$, variational family $q_{\theta}(z)$, number of \gls{MCMC} iterations $t$
		\STATE {\bfseries Output:} variational parameters $\theta$
		\STATE Initialize $\theta$ randomly, initialize $C=0$
		\WHILE{not converged}
			\STATE \verb|# Sample from q:|
			\STATE Sample $z_0\sim q_{\theta}(z_0)$
			\STATE Sample $z\sim Q^{(t)}(z\g z_0)$ (run $t$ \gls{MCMC} steps)
			\STATE \verb|# Estimate the gradient:|
			\STATE Estimate $\widehat{\nabla}_{\theta}\E{q_{\theta}(z_0)}{f_{\theta}(z_0)}$ (\Cref{eq:gradient_term1})
			\STATE Estimate $\widehat{\nabla}_{\theta}\E{q_{\theta}^{(t)}(z)}{f_{\theta}(z)}$ (\Cref{eq:gradient_term2} with control var.)
			\STATE Obtain \hspace*{-1pt} $\widehat{\nabla}_{\theta}\Lcal_{\textrm{VCD}}\! = \!\widehat{\nabla}_{\theta}\mathbb{E}_{q_{\theta}^{(t)}\!(z)}\![f_{\theta}(z)]-\widehat{\nabla}_{\theta}\mathbb{E}_{q_{\theta}\!(z)}\![f_{\theta}(z)]$
			\STATE \verb|# Update the control variate:|
			\STATE Set $C \leftarrow \gamma C + (1-\gamma) f_{\theta}(z)$
			\STATE \verb|# Take gradient step:|
			\STATE Set $\theta \leftarrow \theta - \rho \cdot \widehat{\nabla}_{\theta}\Lcal_{\textrm{VCD}}$
		\ENDWHILE
	\end{algorithmic}
\end{algorithm}

\citet{Zhang2018}  
described a method to combine \gls{MCMC} with \gls{VI}  that tries to directly minimize  
$\textrm{KL}(q_{\theta}^{(t)}(z) \;||\;  p(z\g x) )$, which as discussed 
in \Cref{subsec:refining} is intractable.  \citet{Zhang2018} drop the intractable entropy term from the \gls{ELBO};
again this can result in unstable and inaccurate optimization for a small number of \gls{MCMC} steps $t$.

\citet{Titsias2017} proposed to firstly apply a 
model reparameterization so that the exact marginal likelihood is preserved, i.e.,
$\int p(x,z) d z =  \int p(x, g(\epsilon\prm\theta)) J(\epsilon\prm  \theta) d \epsilon$, where 
$g(\epsilon\prm\theta)$ is a parametrized invertible transformation 
and $J(\epsilon\prm  \theta)$ the determinant of its Jacobian. The method then learns the variational parameters 
$\theta$ by maximizing an \gls{ELBO} under an \gls{MCMC} distribution using an \textsc{em}-like procedure. 
While such an approach can more accurately minimize  
a divergence of the form $\textrm{KL}(q_{\theta}^{(t)}(z) \;||\;  p(z\g x) )$,
it can suffer from the weak gradient problem (see Section \ref{subsec:refining}),
and it is only applicable to differentiable models.
  
Finally, \citet{Habib2019} combine \gls{VI} and \gls{MCMC}
by first fitting a parameterized variational approximation in an augmented space;
this requires to specify variational families for the auxiliary variables.
Instead, we directly operate on the latent variable space.
  
The \gls{VCD} developed in this paper
shares also similarities with contrastive divergence procedures for performing 
\gls{ML} estimation of model parameters in undirected graphical models 
such as restricted Boltzmann machines \citep{Hinton2002}. There, the loss 
function takes a similar discrepancy form between \gls{KL} divergences that involve 
the actual data distribution $p_{\textrm{data}}(x)$ and an \gls{MCMC}-improved
distribution $p^{(t)}(x)$  that converges to the model distribution
$p_{\textrm{model}}(x)$. The fundamental difference with our method 
is that these approaches are designed for model parameter estimation 
while ours is suitable for approximate Bayesian inference.  


\section{Experiments}
\label{sec:experiments}
\glsresetall

Here we demonstrate the algorithm described in \Cref{subsec:full_algorithm}, which minimizes the \gls{VCD} with respect to the variational parameters $\theta$.

In \Cref{subsec:experiments_toy}, we showcase the procedure on a set of toy experiments involving a two-dimensional target distribution. We show that the variational distribution $q_{\theta}(z)$ fitted by minimizing the divergence $\Lcal_{\textrm{VCD}}(\theta)$ has higher variance than the variational distribution fitted by minimizing the standard \gls{KL} divergence.
In \Cref{subsec:experiments_real}, we run experiments on two latent variable models, namely, a matrix factorization model and a \gls{VAE}, using amortized variational distributions. We show that the resulting models fitted with the \gls{VCD} achieve better predictive performance on held-out data.

\subsection{Toy experiments}
\label{subsec:experiments_toy}

To showcase the \gls{VCD}, we approximate a set of synthetic distributions defined on a two-dimensional space: a Gaussian, a mixture of two Gaussians, and a banana distribution. Their densities are given in \Cref{tab:toy_distributions}.

Our goal is to experimentally check that the \gls{VCD} favors higher variance distributions $q_{\theta}(z)$ compared to the standard \gls{KL} divergence used in \gls{VI}. This is because the divergence $\Lcal_{\textrm{VCD}}(\theta)$ converges asymptotically to the symmetrized \gls{KL} divergence between $q_{\theta}(z)$ and the target distribution (see \Cref{subsec:divergence}).

We use two different variational families $q_{\theta}(z)$: a Gaussian distribution and a mixture of two Gaussians. In both cases, the Gaussian components have diagonal covariances. (See Appendix~3 for the mathematical details on minimizing the \gls{VCD} for these specific variational distributions.)

\parhead{Experimental settings.}
Our algorithm of choice to improve the variational distribution is \gls{HMC} \citep{Neal2011}. We set the number of \gls{HMC} iterations $t=3$. We use $5$ leapfrog steps.

We run $20{,}000$ iterations of \Cref{alg:full_algorithm} ($50{,}000$ iterations instead when $q_{\theta}(z)$ is a mixture). We set the learning rate $\eta=0.1$ for the mean parameters, $\eta=0.005$ for the standard deviation, and $\eta=0.001$ for the mixture weights. We additionally decrease the learning rate by a factor of $0.9$ every $2{,}000$ iterations.

\begin{figure*}[t]
	\centering
	\subfigure[\label{subfig:experiments_toy_1}]{\includegraphics[width=0.24\textwidth]{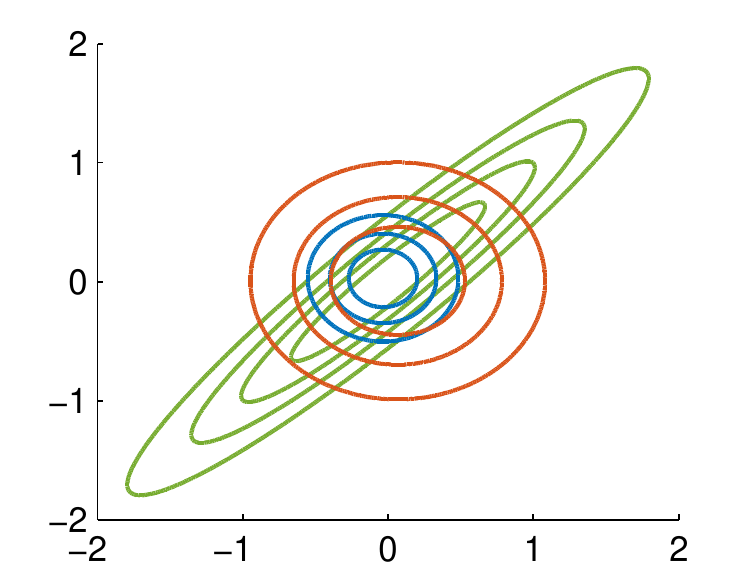}}
	\subfigure[\label{subfig:experiments_toy_2}]{\includegraphics[width=0.24\textwidth]{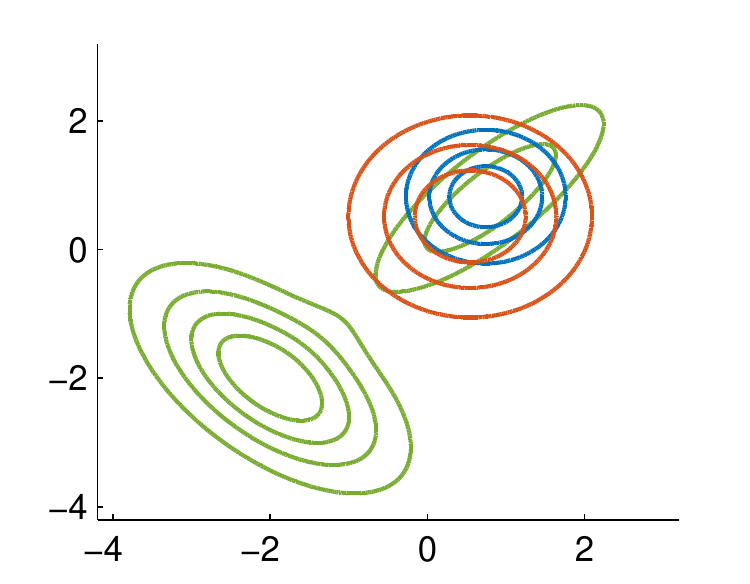}}
	\subfigure[\label{subfig:experiments_toy_3}]{\includegraphics[width=0.24\textwidth]{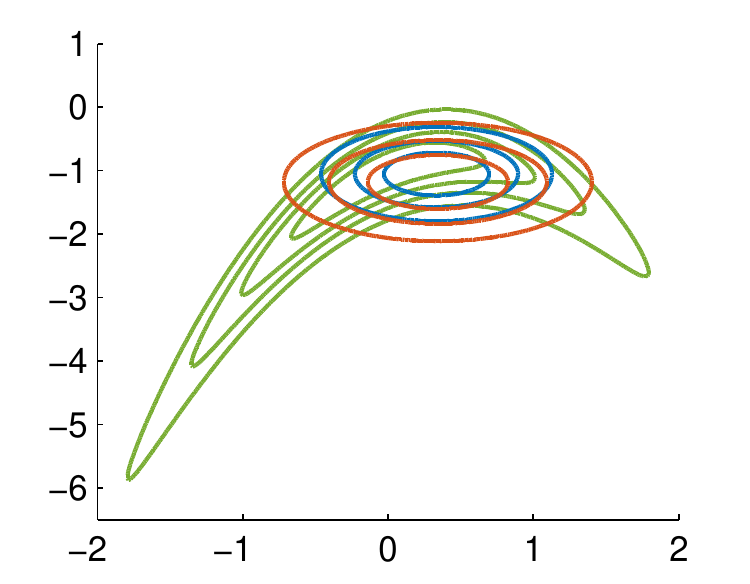}}
	\subfigure[\label{subfig:experiments_toy_4}]{\includegraphics[width=0.24\textwidth]{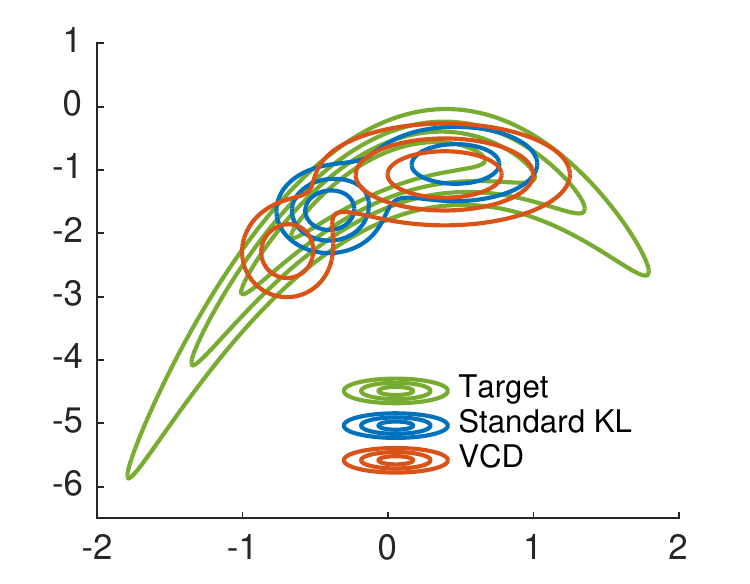}}
	\vspace*{-0.3cm}
	\caption{Examples of fitting variational distributions to several targets (green contours) by applying the standard \gls{KL} divergence (blue contours) and the \gls{VCD} from \Cref{subsec:divergence} (red contours). In the first three panels (a)-(c) the variational distribution is a factorized Gaussian, while in the last panel (d) the variational distribution is a two-component Gaussian mixture. In all cases the \gls{VCD} leads to distributions with higher variance, since $\Lcal_{\textrm{VCD}}(\theta)$ converges asymptotically to the symmetrized \gls{KL} divergence.\label{fig:experiments_toy}} 
\end{figure*}

\parhead{Results.} \Cref{fig:experiments_toy} shows the contour plots of the synthetic target distributions (green), together with the contour plots of the fitted variational distribution $q_{\theta}(z)$. For comparisons, we show the distribution $q_{\theta}(z)$ obtained when optimizing the standard \gls{KL} divergence in \Cref{eq:elbo} (blue), together with the distribution $q_{\theta}(z)$ obtained when optimizing the \gls{VCD} in \Cref{eq:new_divergence_rearranged} (red). Both variational methods were initialized to the same values. In \Cref{subfig:experiments_toy_1,subfig:experiments_toy_2,subfig:experiments_toy_3}, the variational distribution is a factorized Gaussian; in \Cref{subfig:experiments_toy_4} it is a two-component Gaussian mixture.

In all cases, the resulting variational distribution $q_{\theta}(z)$ has higher variance when the objective is the \gls{VCD}. As discussed above, this is due to the asymptotic properties of the divergence; specifically, the \gls{VCD} eventually converges to the symmetrized \gls{KL} divergence between $q_{\theta}(z)$ and the target. This effect is apparent despite the small number of \gls{HMC} iterations ($t=3$) because the target is a simple two-dimensional distribution.

\begin{table}[t]
	\centering
 	\small
	\caption{Synthetic distributions used in the toy experiments.\label{tab:toy_distributions}}
	{\setlength{\tabcolsep}{1.2pt}
	\begin{tabular}{cc} \toprule
		name & $p(z)$ \\ \midrule
		Gaussian & {\scriptsize$ \Ncal\left(z\;\bigg|\; \left[\begin{array}{c} 0 \\ 0 \end{array}\right], \left[\! \begin{array}{cc} 1 & 0.95  \\ 0.95 & 1 \end{array} \!\right]\right) $} \vspace*{5pt} \\
		mixture & {\scriptsize $0.3\Ncal\!\left(\! z \bigg| \!\left[\! \arraycolsep=1.2pt \begin{array}{c} 0.8  \\ 0.8 \end{array} \!\right]\!,\! \left[\! \arraycolsep=1.2pt \begin{array}{cc} 1 & 0.8  \\ 0.8 & 1 \end{array} \!\right]\right)\!\! +\! 0.7\Ncal\!\left(\! z\bigg|\! \left[\! \arraycolsep=1.2pt \begin{array}{c} -2  \\ -2 \end{array} \!\right]\!,\! \left[\! \arraycolsep=0.9pt \begin{array}{cc} 1 & -0.6 \\ -0.6 & 1 \end{array}\!\right]\right)$ } \vspace*{5pt} \\
		banana & {\scriptsize
				$ \Ncal\left( \left[\begin{array}{c} z_1 \\ z_2+z_1^2+1 \end{array}\right] \bigg| \left[\begin{array}{c} 0 \\ 0 \end{array}\right], \left[\begin{array}{cc} 1 & 0.9 \\ 0.9 & 1 \end{array}\right] \right)$} \vspace*{5pt} \\ \bottomrule
	\end{tabular} }
\end{table}

\subsection{Experiments on latent variable models}
\label{subsec:experiments_real}

We now consider latent variable models, where there is a local latent variable $z_n$ for each observation $x_n$ in the dataset, and the joint distribution factorizes as $p_{\phi}(x, z)=\prod_n p(z_n) p_{\phi}(x_n\g z_n)$. Here, $\phi$ is a model parameter that we wish to learn via \gls{ML}.

We use amortized \gls{VI} for inference over the latent variables, i.e., the distribution $q_{\theta}(z_n)\equiv q_{\theta}(z_n\g x_n)$. The improved variational distribution, after running $t$ \gls{HMC} steps, is $q_{\theta}^{(t)}(z_n\g x_n)=\int Q_n^{(t)}(z_n\g z_n^{(0)}) q_{\theta}(z_n^{(0)} \g x_n) dz_n^{(0)}$, where the $t$-step \gls{HMC} kernel $Q_n^{(t)}(z_n\g z_n^{(0)})$ has the posterior $p_{\phi}(z_n\g x_n)$ as its stationary distribution.

Our goal is to show empirically that the fitted models lead to better predictive performance when we optimize the variational parameters $\theta$ by minimizing the \gls{VCD}, compared to the standard \gls{VI} setting.

\parhead{Models.}
We consider two models. The first one is Bayesian logistic matrix factorization, a model of binary data. The likelihood $p_{\phi}(x_n\g z_n)=\prod_d p_{\phi}(x_{nd}\g z_n)$ is a product of Bernoulli distributions, each with parameter $\textrm{sigmoid}(z_n^\top \phi_d + \phi_d^{(0)})$, where $\textrm{sigmoid}(x)=1/(1+e^{-x})$. The model parameters are the weights $\phi_d$ and intercepts $\phi_d^{(0)}$ for each dimension $d$.

The second model is a \gls{VAE} \citep{Kingma2014}, which uses a density network \citep{MacKay1995} to define the likelihood $p_{\phi}(x_n\g z_n)$. That is, the likelihood is parameterized by a neural network with parameters $\phi$. We choose a fully connected neural network with two hidden layers of $200$ hidden units each and ReLu activation functions. We use a Bernoulli likelihood $p_{\phi}(x_n\g z_n)$, similarly to Bayesian logistic matrix factorization, and thus the output layer of the neural network performs a sigmoid transformation. 

For both models, we consider a standard multivariate Gaussian prior $p(z_n)=\Ncal(z_n\g 0,I)$.

\parhead{Methods.}
We compare three different objectives to perform the optimization: standard \gls{KL}, the method of \citet{Hoffman2017}, and the divergence $\Lcal_{\textrm{VCD}}(\theta)$. The standard \gls{KL} objective involves an explicit variational distribution. Both the objective of \citet{Hoffman2017} and the \gls{VCD} divergence involve an improved \gls{HMC} distribution $q_{\theta}^{(t)}(z_n\g x_n)$.

We fit the model parameters $\phi$ via \gls{ML} by maximizing the objective $\sum_n \E{q_{\theta}^{(t)}(z_n\g x_n)}{\log p_{\phi}(x_n\g z_n)}$. The variational distribution $q_{\theta}^{(t)}(z_n\g x_n)=q_{\theta}(z_n\g x_n)$ for the standard \gls{KL} method, as there is no improved distribution. For the other two methods, $q_{\theta}^{(t)}(z_n\g x_n)$ is the distribution improved with \gls{HMC}, and maximizing the objective with respect to $\phi$ corresponds essentially to a Monte Carlo expectation maximization algorithm \citep{Wei1990}.

We fit the variational parameters $\theta$ by optimizing the corresponding divergence, according to the method. Both the standard \gls{KL} and the method of \citet{Hoffman2017} optimize the \gls{ELBO} in \Cref{eq:elbo}. Thus, the method of \citet{Hoffman2017} does not incorporate feedback from the \gls{HMC} chain into learning $\theta$, as discussed in \Cref{sec:related}. Instead, the \gls{VCD} method optimizes $\Lcal_{\textrm{VCD}}(\theta)$ in \Cref{eq:new_divergence}.

\parhead{Datasets.}
We use two datasets. The first one is the binarized \textsc{mnist} data \citep{Salakhutdinov2008_quantitative}, which contains $50{,}000$ training images and $10{,}000$ test images of hand-written digits. The second dataset is Fashion-\textsc{mnist} \citep{Xiao2017}, which contains $60{,}000$ training images and $10{,}000$ test images of clothing items. We binarize the Fashion-\textsc{mnist} images with a threshold at $0.5$. Images in both datasets are of size $28\times 28$ pixels.


\begin{table}[t]
	\centering
 	\small
 	\vspace*{-0.25cm}
	\caption{Marginal log-likelihood on the test set. Fitting an improved distribution by minimizing the \acrshort{VCD} leads to the best predictive performance.\label{tab:results_llh}}
	\vspace*{4pt}
	\begin{tabular}{ccc} \toprule
		& \multicolumn{2}{c}{average test log-likelihood} \\
		method & \textsc{mnist} & Fashion-\textsc{mnist} \\ \midrule
		Standard \acrshort{KL} & $-111.20$ & $-127.43$ \\
		\citet{Hoffman2017} & $-103.61$ & $-121.86$ \\
		$\acrshort{VCD}$ (this paper) & $\mathbf{-101.26}$ & $\mathbf{-121.11}$ \\ \bottomrule
	\end{tabular}\\
	{(a) Bayesian logistic matrix factorization.} \vspace*{6pt}\\
	\begin{tabular}{ccc} \toprule
		& \multicolumn{2}{c}{average test log-likelihood} \\
		method & \textsc{mnist} & Fashion-\textsc{mnist} \\ \midrule
		Standard \acrshort{KL} & $-98.46$ & $-124.63$ \\
		\citet{Hoffman2017} & $-96.23$ & $-117.74$ \\
		$\acrshort{VCD}$ (this paper) & $\mathbf{-95.86}$ & $\mathbf{-117.65}$ \\ \bottomrule
	\end{tabular}\\
	{(b) \Acrlong{VAE}.}
 	\vspace*{-0.25cm}
\end{table}

\parhead{Variational family.}
The variational family is a Gaussian, $q_{\theta}(z_n\g x_n)=\Ncal(z_n\g \mu_{\theta}(x_n), \Sigma_{\theta}(x_n))$, whose mean and covariance are parameterized using two separate fully connected neural networks with two hidden layers of $200$ units each. The neural networks have ReLu units, and the covariance $\Sigma_{\theta}(x_n)$ is set to be diagonal. The neural network for the covariance has non-linear activations in the output layer to ensure positive outputs. In particular, for each entry corresponding to the standard deviation, the neural network activation function is a modified softplus, $\textrm{softplus}(x)=\log(\exp\{10^{-4}\}+\exp\{x\})$. The modified softplus avoids numerical issues; it ensures that the variances are above a small threshold at $10^{-8}$.

\parhead{Experimental settings.}
We set the number of \gls{HMC} iterations $t=8$, using $5$ leapfrog steps. We set the learning rate $\eta=5\times 10^{-4}$ for the variational parameters corresponding to the mean, $\eta=2.5\times 10^{-4}$ for the variational parameters corresponding to the covariance, and $\eta=5\times 10^{-4}$ for the model parameters $\phi$.
We additionally decrease the learning rate by a factor of $0.9$ every $15{,}000$ iterations.
We set the dimensionality of $z_n$ to $50$ for Bayesian logistic matrix factorization, and to $10$ for the \gls{VAE}.

We run $400{,}000$ iterations of each optimization algorithm. We perform stochastic \gls{VI} by subsampling a minibatch of observations at each iteration \citep{Hoffman2013}; we set the minibatch size to $100$.

For the \gls{VCD}, the control variates are local, i.e., there is a $C_n$ for each datapoint. However, in the earlier iterations of the optimization procedure we set the control variates to the same global value, $C_n=C$. The reason is that in these earlier iterations the model is highly non-stationary, as the model parameters $\phi$ change more significantly. We found that introducing local control variates at the beginning may lead to unstabilities; therefore we only introduce the local control variates $C_n$ after $3{,}000$ iterations. Before that, we update the global control variate $C$ taking the mean of the stochastic estimates in the minibatch. After iteration $3{,}000$, we let each $C_n$ be updated independently.

\parhead{Evaluation.}
We compute the average marginal test log-likelihood. 
For each test datapoint $x_n^\star$, we estimate the marginal log-likelihood using importance sampling,
\begin{equation}\label{eq:approx_log_lik}
	\log p_{\phi}(x_n^\star) \approx \log \frac{1}{S} \sum_{s=1}^{S} \frac{p_{\phi}(x_n^\star\g z_n^{(s)})p(z_n^{(s)})}{r_{\theta}(z_n^{(s)}\g x_n^\star)},
\end{equation}
where $z_n^{(s)} \sim r_{\theta}(z_n\g x_n^\star)$. For each test instance $x_n^\star$,
we use three different proposals $r_{\theta}(z_n\g x_n^\star)$ and keep the highest resulting value (note that the approximation in \Cref{eq:approx_log_lik} gives a lower bound of the marginal log-likelihood).
The first proposal is an overdispersed version of the amortized variational distribution $q_{\theta}(z_n\g x_n^\star)$. It is a Gaussian distribution whose mean is equal to the mean of $q_{\theta}(z_n\g x_n^\star)$ but its standard deviation is $1.2$ times larger. The second proposal is a Gaussian distribution whose mean is set to the mean of the samples resulting after running an \gls{HMC} chain initialized at $q_{\theta}(z_n\g x_n^\star)$ (we run the chain for $600$ iterations and obtain the mean of the last $300$ samples). We set the standard deviation of the proposal $1.2$ times larger than the standard deviation of $q_{\theta}(z_n\g x_n^\star)$. The third proposal is similar to the second one, but we set its standard deviation $1.2$ times larger than the standard deviation of the last $300$ \gls{HMC} samples. In all cases, we set $S=20{,}000$ samples.


\begin{figure}[t]
	\centering
	\subfigure[\textsc{mnist}.]{\includegraphics[width=0.24\textwidth]{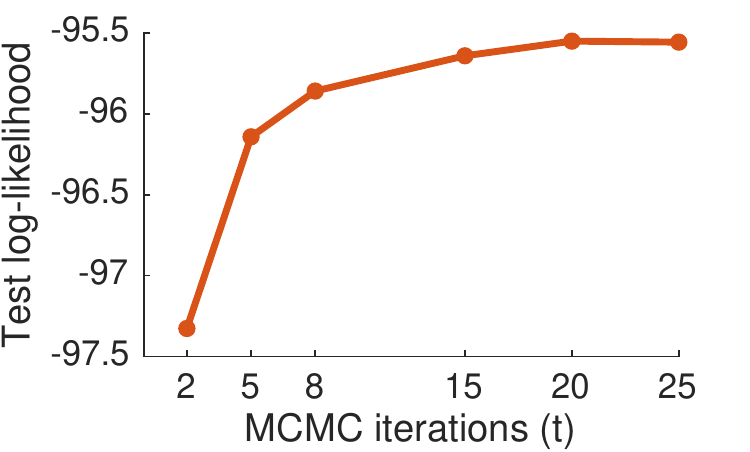}}\hspace*{-0pt}
	\subfigure[Fashion-\textsc{mnist}.]{\includegraphics[width=0.24\textwidth]{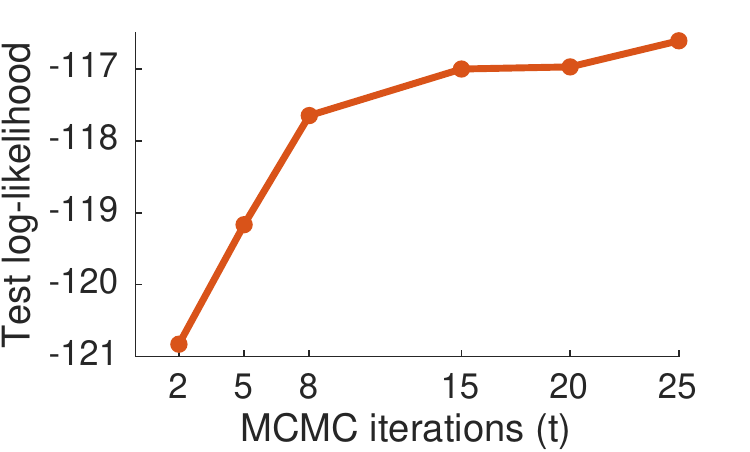}}
	\vspace*{-0.4cm}
	\caption{Estimates of the marginal log-likelihood on the test set for the \acrshort{VAE} as a function of the number of \acrshort{MCMC} steps $t$. Increasing the number of \acrshort{MCMC} steps improves the performance.}
	\label{fig:results_vae_vary_t}
	\subfigure[\textsc{mnist}.]{\includegraphics[width=0.24\textwidth]{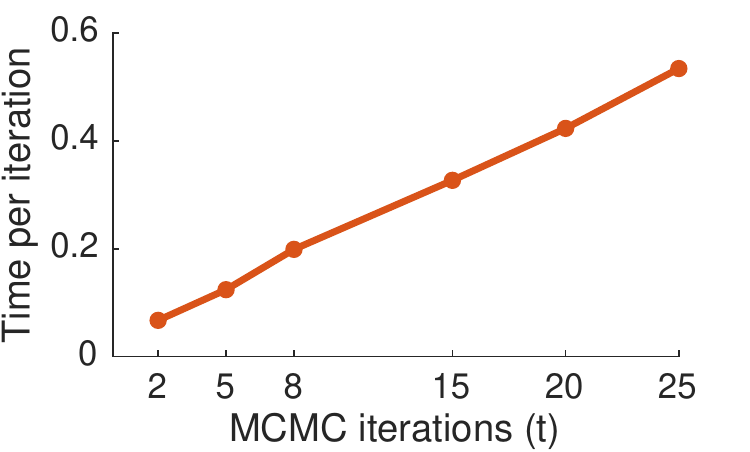}}\hspace*{-0pt}
	\subfigure[Fashion-\textsc{mnist}.]{\includegraphics[width=0.24\textwidth]{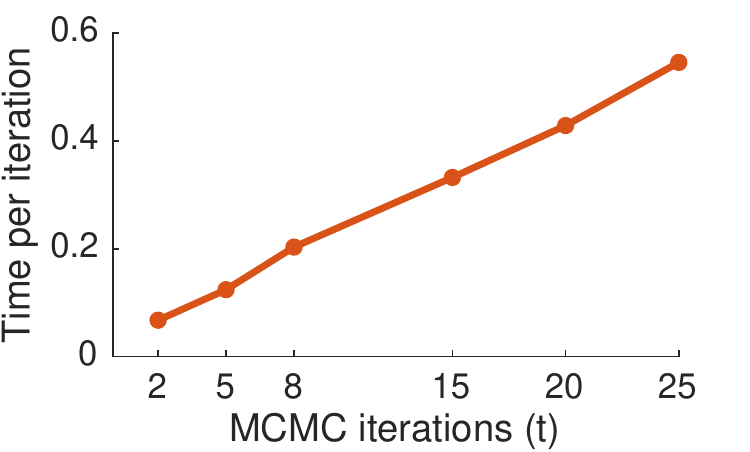}}
	\vspace*{-0.4cm}
	\caption{Average time (in seconds) per iteration of the inference algorithm for the \acrshort{VAE} as a function of the number of \acrshort{MCMC} steps.}
	\label{fig:results_vae_vary_t_telapsed}
	\vspace*{-0.2cm}
\end{figure}

\parhead{Results.}
\Cref{tab:results_llh} displays the average test log-likelihood. Optimizing the \gls{VCD} leads to the best results for all methods and datasets. 
We can conclude that optimizing the \gls{VCD} is in general the best approach, since it is at least as good as the state of the art. It outperforms standard amortized \gls{VI} because it refines the variational distribution with \gls{HMC}, and it outperforms the method of \citet{Hoffman2017} because it uses feedback from the \gls{HMC} samples when fitting the variational parameters $\theta$.

\Cref{tab:results_llh} was obtained with $t=8$ \gls{HMC} steps. We now study the impact of $t$ on the results. \Cref{fig:results_vae_vary_t} shows the test log-likelihood for the \gls{VAE} as a function of the number of \gls{HMC} steps, ranging from $t=2$ to $t=25$. As expected, increasing the number of steps improves the performance. Even for a small value $t=2$, the test log-likelihood is better than for the standard \gls{KL} method (see \Cref{tab:results_llh}b).

The improvement comes at the cost of computational complexity. \Cref{fig:results_vae_vary_t_telapsed} shows that the average time per iteration for fitting the \gls{VAE} by minimizing the \gls{VCD} increases linearly with the number of \gls{HMC} steps. (No parallelism or \textsc{gpu} acceleration was used.) 
We also found that the optimization of the \gls{VCD} is slightly faster than the method of \citet{Hoffman2017} for all models and datasets. This is counter-intuitive, as the \gls{VCD} requires a few additional computations, although their computational complexity is negligible compared to the \gls{HMC} steps. Therefore, both methods should run roughly equally as fast. We believe that the differences we observed are implementation-specific.

To sum up, more computation leads to better results, but even a few \gls{HMC} steps are advantageous compared to the minimization of the standard \gls{KL} divergence.

\section{Conclusion}
\label{sec:conclusion}
\glsresetall

We have proposed a method to improve the approximating distribution in \gls{VI} by running a \gls{MCMC} algorithm that targets the posterior of a probabilistic model. This leads to an implicit approximating distribution with an intractable density, which makes it challenging to minimize the \gls{KL} divergence between the approximation and the posterior. To address that, we have developed a divergence, called \gls{VCD}, that can be tractably optimized with respect to the variational parameters; in particular, we can form unbiased Monte Carlo estimators of its gradient. The \gls{VCD} differs from the standard \gls{KL} in that it measures the discrepancy between the improved and the initial variational distributions. We have shown empirically that minimizing the \gls{VCD} leads to better predictive performance in latent variable models.

One line for future research is to use the \gls{VCD} to design tests for assessing the quality (or exactness) of a given variational distribution, similarly to the goal of \citet{Yao2018}. For that, we can use a property of the \gls{VCD}---that it allows us to obtain unbiased estimates of the actual value of the divergence (based on \Cref{eq:new_divergence_rearranged}). Such tests can rely on the \gls{VCD} being a proper divergence, as it takes non-negative values and it becomes zero only when the variational approximation matches the exact posterior.

\section*{Acknowledgements}
Francisco J.\ R.\ Ruiz is supported by the EU Horizon 2020 programme (Marie Sk\l{}odowska-Curie Individual Fellowship, grant agreement 706760). The authors thank Michael Figurnov for his useful comments.


\bibliography{fjrrLibrary}
\bibliographystyle{icml2019}

\newpage

\appendix

\section{Simplification of the \acrshort{VCD}}

Here we show how to express the simplified expression of the \gls{VCD} as the difference of two expectations. We start from the definition in terms of the \gls{KL} divergences,
\begin{equation}\label{eq2:new_divergence}
	\begin{split}
		& \Lcal_{\textrm{VCD}}(\theta) \\
		& = \Lcal_{\textrm{diff}}(\theta) + \textrm{KL}(q_{\theta}^{(t)}(z) \;||\; q_{\theta}(z)) \\
		& = \textrm{KL}(q_{\theta}(z) \;||\; p(z\g x)) - \textrm{KL}(q_{\theta}^{(t)}(z) \;||\; p(z\g x)) \\
		&\quad + \textrm{KL}(q_{\theta}^{(t)}(z) \;||\; q_{\theta}(z)).
	\end{split}
\end{equation}
We now apply the definition of the \gls{KL} divergence,
\begin{equation}
	\begin{split}
		& \Lcal_{\textrm{VCD}}(\theta) \\
		& = \E{q_{\theta}(z)}{\log\frac{q_{\theta}(z)}{p(z\g x)}} - \E{q_{\theta}^{(t)}(z)}{\log\frac{q_{\theta}^{(t)}(z)}{p(z\g x)}} \\
		&\quad + \E{q_{\theta}^{(t)}(z)}{\log\frac{q_{\theta}^{(t)}(z)}{q_{\theta}(z)}}.
	\end{split}
\end{equation}
Next we expand the logarithms. The expectation of the log-density $\log q_{\theta}^{(t)}(z)$ cancels out because it appears with different signs in the second and third terms. We also rewrite the posterior $p(z\g x)=p(x,z)/p(x)$,
\begin{equation}
	\begin{split}
		& \Lcal_{\textrm{VCD}}(\theta) \\
		& = \E{q_{\theta}(z)}{\log\frac{q_{\theta}(z)p(x)}{p(x,z)}} - \E{q_{\theta}^{(t)}(z)}{\log\frac{q_{\theta}(z)p(x)}{p(x,z)}}.
	\end{split}
\end{equation}
The marginal log-likelihood $\log p(x)$ does not depend on $z$ and can be taken out of the expectation. Since it appears with different signs, it cancels out. We now recognize that the argument of each expectation---after having canceled out the marginal log-likelihood---is the negative instantaneous \gls{ELBO}, defined as
\begin{equation}\label{eq2:instantaneous_elbo}
	f_{\theta}(z) \triangleq \log p(x,z) - \log q_{\theta}(z).
\end{equation}
Thus, we finally have
\begin{equation}\label{eq2:new_divergence_rearranged}
	\Lcal_{\textrm{VCD}}(\theta) = -\E{q_{\theta}(z)}{f_{\theta}(z)} + \E{q_{\theta}^{(t)}(z)}{f_{\theta}(z)}.
\end{equation}

\section{Generalization of the \acrshort{VCD}}

The \gls{VCD} divergence can be generalized with a parameter $\alpha$ that downweights the two \gls{KL} terms involving the improved distribution $q_{\theta}^{(t)}(z)$. More in detail, we define the $\alpha$-generalized \gls{VCD} as
\begin{eqnarray}
	&& \Lcal_{\textrm{VCD}}^{(\alpha)}(\theta) = \textrm{KL}(q_{\theta}(z) \;||\; p(z\g x)) \\
	&& + \alpha\left[\textrm{KL}(q_{\theta}^{(t)}(z) \;||\; q_{\theta}(z)) - \textrm{KL}(q_{\theta}^{(t)}(z) \;||\; p(z\g x)) \right]. \nonumber
\end{eqnarray}
For any $0\leq \alpha\leq 1$, the $\alpha$-generalized \gls{VCD} is also a proper divergence because it satisfies the two desired criteria---it is non-negative and it becomes zero only when $q_{\theta}=p(z\g x)$. Moreover, it leads to tractable optimization because the intractable log-density $\log q_{\theta}^{(t)}(z)$ also cancels out in this expression.

By varying $\alpha$, the $\alpha$-generalized \gls{VCD} interpolates between the standard \gls{KL} divergence of \gls{VI} (for $\alpha=0$) and the \gls{VCD} in \Cref{eq2:new_divergence} (for $\alpha=1$). When the number of \gls{MCMC} steps is large, this is effectively an interpolation between the standard \gls{KL} and the symmetrized \gls{KL} divergence.

The $\alpha$-generalized \gls{VCD} is useful when the \gls{MCMC} method does not mix well. To see this, consider that due to slow mixing, the improved distribution $q_{\theta}^{(t)}(z)\approx q_{\theta}(z)$. In this case, the divergence $\Lcal_{\textrm{VCD}}(\theta)\approx 0$ for any value of the variational parameters, but the $\alpha$-generalized \gls{VCD} becomes proportional to the standard \gls{KL}, $\Lcal_{\textrm{VCD}}^{(\alpha)}(\theta)\approx (1-\alpha)\textrm{KL}(q_{\theta}(z) \;||\; p(z\g x))$. Therefore, the $\alpha$-generalized \gls{VCD} may lead to more robust optimization when the \gls{MCMC} method does not mix well.

In our experiments, we consider the non-generalized \gls{VCD} in \Cref{eq2:new_divergence} because we did not find any mixing issues with our \gls{MCMC} method. The derivations of the gradients in the main paper can be straightforwardly generalized for the case where the objective is $\Lcal_{\textrm{VCD}}^{(\alpha)}(\theta)$.

\section{Particularizations of the Gradients}

Here we derive the gradients of the \gls{VCD} for two choices of the variational distribution $q_{\theta}(z)$, namely, a Gaussian and a mixture of Gaussians.

\subsection{Gaussian Variational Distribution}

We now show how to obtain the gradient of the \gls{VCD} in the case where the distribution $q_{\theta}(z)$ is Gaussian.

Consider $q_{\theta}(z)=\Ncal(z\g \mu, \Sigma)$, i.e., a Gaussian distribution with mean $\mu$ and covariance $\Sigma$. That is,
\begin{equation}\label{eq2:gaussian_pdf}
	\begin{split}
		\log q_{\theta}(z) = & -\frac{D}{2}\log(2\pi) - \frac{1}{2}\log|\Sigma| \\
		& -\frac{1}{2}(z-\mu)^\top \Sigma^{-1}(z-\mu)
	\end{split}
\end{equation}
Here, $\theta=[\mu,\Sigma]$ denotes the variational parameters, and $D$ is the dimensionality of the latent variable, $z\in\Reals^D$.

The definition of the \gls{VCD} is in \Cref{eq2:new_divergence_rearranged}.
Since it consists of a difference of two expectations of the same function $f_{\theta}(z)$, the terms that are constant with respect to $z$ in $f_{\theta}(z)$ cancel out. Specifically, the term $-\frac{D}{2}\log(2\pi) - \frac{1}{2}\log|\Sigma|$ from \Cref{eq2:gaussian_pdf}, which appears in $f_{\theta}(z)$ (\Cref{eq2:instantaneous_elbo}), is constant with respect to $z$; therefore it cancels out. This leads to the simplified objective
\begin{equation}\label{eq2:objective_somewhat_simplified}
	 \Lcal_{\textrm{VCD}}(\theta) = -\E{q_{\theta}(z)}{g_{\theta}(z)} + \E{q_{\theta}^{(t)}(z)}{g_{\theta}(z)},
\end{equation}
where we have introduced the shorthand notation $g_{\theta}(z)$ for the (simplified) argument of the expectation,
\begin{equation}
	g_{\theta}(z)\triangleq \log p(x,z) + \frac{1}{2}(z-\mu)^\top \Sigma^{-1}(z-\mu).
\end{equation}
\Cref{eq2:objective_somewhat_simplified} can be further simplified by computing the exact expectation of the quadratic form that appears in the first expectation,
\begin{equation}\label{eq2:objective_simplified}
	\Lcal_{\textrm{VCD}}(\theta) = -\E{q_{\theta}(z)}{\log p(x,z)} - \frac{D}{2} 
	+ \E{q_{\theta}^{(t)}(z)}{g_{\theta}(z)}.
\end{equation}

These expressions are also valid when using amortized inference with a Gaussian variational distribution. That is, when $\mu$ and $\Sigma$ are functions of the data $x$, $\mu=\mu_{\theta}(x)$ and $\Sigma=\Sigma_{\theta}(x)$.

\parhead{Taking the gradients.}
We now derive the expressions for the gradient of the \gls{VCD} with respect to the variational parameters.
We parameterize the Gaussian in terms of its mean and the Cholesky decomposition of its covariance. That is,
the variational parameters are $\mu$ and $L$, where $L$ is a lower triangular matrix such that $LL^\top=\Sigma$.
The reparameterization transformation in terms of a standard Gaussian $q(\varepsilon)=\Ncal(\varepsilon\g 0, I)$
is given as $\varepsilon\sim q(\varepsilon)$, $z=h_{\theta}(\varepsilon)=\mu+L\varepsilon$.


The gradient of the (negative) first term in \Cref{eq2:objective_simplified} is directly given by the reparameterization gradient,
\begin{equation}
	\begin{split}
		& \nabla_{\theta} \E{q_{\theta}(z)}{\log p(x,z)} \\
		& = \E{q(\varepsilon)}{ \nabla_z \log p(x,z) \big|_{z=h_{\theta}(\varepsilon)} \times \nabla_{\theta} h_{\theta}(\varepsilon)},
	\end{split}
\end{equation}
where $\theta=[\mu,L]$ denotes the variational parameters.

For the second expectation in \Cref{eq2:objective_simplified}, we apply the derivation in the main paper,
\begin{equation}
	\begin{split}
		& \nabla_{\theta} \E{q_{\theta}^{(t)}(z)}{g_{\theta}(z)} = \E{q_{\theta}^{(t)}(z)}{\nabla_{\theta} g_{\theta}(z)} \\
		& +\E{Q^{(t)}(z\g z_0) q_{\theta}(z_0)}{g_{\theta}(z) \times \nabla_{\theta} \log q_{\theta}(z_0)}.
	\end{split}
\end{equation}
Note that the gradient $\nabla_{\theta} g_{\theta}(z)$ only involves the gradient of the quadratic form, since the model $p(x,z)$ does not depend on $\theta$. The gradient $\nabla_{\theta} \log q_{\theta}(z_0)$ is the gradient of the Gaussian log-density,
\begin{equation}
	\begin{split}
		& \nabla_\mu \log q_{\theta}(z_0) = L^{-\top}L^{-1}(z_0-\mu),\\
		& \nabla_{L} \log q_{\theta}(z_0) \\
		& = -\Omega + \left( L^{-\top}L^{-1}(z_0\!-\!\mu)(z_0\!-\!\mu)^\top\! L^{-\top}\right) \odot M,\nonumber
	\end{split}
\end{equation}
where $\Omega$ is a diagonal matrix whose entries are given by the element-wise inverse of the diagonal entries of $L$, the symbol $\odot$ denotes the element-wise product, and $M$ is a lower triangular masking matrix of ones (it contains zeros above the main diagonal).

The above expressions are also valid when the variational distribution is a fully factorized Gaussian, in which case $L$ is a diagonal matrix whose entries correspond to the standard deviation of each component. This is the setting that we consider in the paper.

\subsection{Mixture of Gaussians}

Consider now that the variational distribution $q_{\theta}(z)$ is a mixture of $K$ components,
\begin{equation}\label{eq2:mixture_pdf}
	q_{\theta}(z) = \sum_{k=1}^{K} w_k q_{\theta_k}(z).
\end{equation}
where each $q_{\theta_k}(z)$ is a Gaussian,
\begin{equation}\label{eq2:mixture_component_pdf}
	q_{\theta_k}(z) = \frac{1}{\sqrt{(2\pi)^D |\Sigma_k|}} e^{ -\frac{1}{2}(z-\mu_k)^\top \Sigma_k^{-1}(z-\mu_k) }.
\end{equation}
The variational parameters are $\theta_k=[\mu_k,L_k]$ for each component, where $L_k$ is the Cholesky decomposition of $\Sigma_k$, i.e., $L_kL_k^\top=\Sigma_k$, as well as the mixture weights $w_k$.

The \gls{VCD} objective is given in \Cref{eq2:new_divergence_rearranged}; however in this case there are no constant terms in $f_{\theta}(z)$ that cancel out as in the Gaussian case. Thus, we obtain the gradient of the \gls{VCD} by computing the gradients $\nabla_{\theta} \E{q_{\theta}(z)}{f_{\theta}(z)}$ and $\nabla_{\theta} \E{q_{\theta}^{(t)}(z)}{f_{\theta}(z)}$ separately.

\parhead{Taking the gradient of the first term.}
We first rewrite the standard \gls{ELBO} term as a sum over the mixture components,
\begin{equation}
	\E{q_{\theta}(z)}{f_{\theta}(z)} = \sum_{k=1}^{K} w_k \E{q_{\theta_k}(z)}{f_{\theta}(z)}.
\end{equation}
We next reparameterize each component, with $\varepsilon\sim q(\varepsilon)=\Ncal(\varepsilon\prm 0,I)$ and $z=h_{\theta_k}(\varepsilon)=\mu_k+L_k\varepsilon$. We rewrite the \gls{ELBO} in terms of this reparameterization,
\begin{equation}
	\E{q_{\theta}(z)}{f_{\theta}(z)} = \sum_{k=1}^{K} w_k \E{q(\varepsilon)}{f_{\theta}(z)\big|_{z=h_{\theta_k}(\varepsilon)}}.
\end{equation}
We now take the gradient of the \gls{ELBO}. The gradient with respect to each component $\theta_k$ is
\begin{equation}\label{eq2:mixture_grad_elbo_thetak}
	\begin{split}
	& \nabla_{\theta_k} \mathbb{E}_{q_{\theta}(z)}\left[f_{\theta}(z)\right] \\
	& = \sum_{k^\prime=1}^{K} w_{k^\prime} \mathbb{E}_{q(\varepsilon)}\left[\nabla_z f_{\theta}(z)\big|_{z=h_{\theta_{k^\prime}}(\varepsilon)}\nabla_{\theta_k} h_{\theta_{k^\prime}}(\varepsilon) \right] \\
	& \quad + \sum_{k^\prime=1}^{K} w_{k^\prime} \mathbb{E}_{q_{\theta_{k^\prime}}(z)}\left[\nabla_{\theta_k} f_{\theta}(z)\right] \\
	& = w_{k} \mathbb{E}_{q(\varepsilon)}\left[\nabla_z f_{\theta}(z)\big|_{z=h_{\theta_{k}}(\varepsilon)}\nabla_{\theta_k} h_{\theta_{k}}(\varepsilon)\right] \\
	& \quad + \mathbb{E}_{q_{\theta}(z)}\left[-\nabla_{\theta_k} \log q_{\theta}(z)\right]\\
	& = w_{k} \mathbb{E}_{q(\varepsilon)}\left[\nabla_z f_{\theta}(z)\big|_{z=h_{\theta_{k}}(\varepsilon)}\nabla_{\theta_k} h_{\theta_{k}}(\varepsilon)\right].
	\end{split}
\end{equation}
We now obtain the gradient of the \gls{ELBO} w.r.t.\ the mixture weights. We build a score function estimator,
\begin{equation}\label{eq2:mixture_grad_elbo_wk}
	\begin{split}
		& \nabla_{w_k} \mathbb{E}_{q_{\theta}(z)}\left[ f_{\theta}(z) \right] \\
		& = \E{q_{\theta}(z)}{f_{\theta}(z)\nabla_{w_k} \log q_{\theta}(z)}  \\
		& = \E{q_{\theta}(z)}{f_{\theta}(z) \frac{1}{q_{\theta}(z)} q_{\theta_k}(z)}  \\
		& = \mathbb{E}_{q_{\theta_k}(z)}\left[ f_{\theta}(z) \right].
	\end{split}
\end{equation}

To sum up, we have obtained a reparameterization gradient for the parameters $\theta_k$ (\Cref{eq2:mixture_grad_elbo_thetak}) and a score function gradient for the mixture weights $w_k$ (\Cref{eq2:mixture_grad_elbo_wk}).

For the reparameterization gradient, one of the quantities that we need to evaluate is the gradient $\nabla_z \log q_{\theta}(z)$. We compute this gradient in a numerically stable manner using the log-derivative trick,
\begin{equation}
	\begin{split}
		& \nabla_z \log q_{\theta}(z) = \frac{1}{q_{\theta}(z)} \sum_{k=1}^{K} w_k \nabla_z q_{\theta_k}(z) \\
		& = \frac{1}{q_{\theta}(z)} \sum_{k=1}^{K} w_k q_{\theta_k}(z) \nabla_z \log q_{\theta_k}(z) \\
		& = \sum_{k=1}^{K} q_{\theta}(k\g z) \nabla_z \log q_{\theta_k}(z),
	\end{split}
\end{equation}
where $q_{\theta}(k\g z)$ is the ``posterior probability'' (under the variational model) of component $k$ given $z$, i.e.,
\begin{equation}
	q_{\theta}(k\g z) \propto \exp\left\{ \log w_k + \log q_{\theta_k}(z) \right\}.
\end{equation}

\parhead{Taking the gradient of the second term.}
The second term in th \gls{VCD} of \Cref{eq2:new_divergence_rearranged} involves an expectation with respect to the improved distribution $q_{\theta}^{(t)}(z)$. As derived in the main paper, the gradient of the second term can in turn be split into two terms. We first obtain the gradient w.r.t.\ the parameters $\theta_k$ of each Gaussian component,
\begin{equation}
	\begin{split}
		& \nabla_{\theta_k} \E{q_{\theta}^{(t)}(z)}{f_{\theta}(z)} = \E{q_{\theta}^{(t)}(z)}{-\nabla_{\theta_k}\log q_{\theta}(z)} \\
		& + \E{q_{\theta}(z_0)}{w_{\theta}(z_0) \nabla_{\theta_k} \log q_{\theta}(z_0)},
	\end{split}
\end{equation}
where we have defined
\begin{equation}
	w_{\theta}(z_0)\triangleq \E{Q^{(t)}(z\g z_0)}{f_{\theta}(z)}.
\end{equation}

Now we make use of the definition of $q_{\theta}(z)$ and substitute \Cref{eq2:mixture_pdf} in the two expressions above, yielding
\begin{equation}\label{eq2:mixture_grad_term2_thetak}
	\begin{split}
		& \nabla_{\theta_k} \E{q_{\theta}^{(t)}(z)}{f_{\theta}(z)} \\
		& = \sum_{k=1}^{K} w_k \E{q_{\theta_k}(z_0)}{\E{Q^{(t)}(z\g z_0)}{-\nabla_{\theta_k}\log q_{\theta}(z)}} \\
		& \quad + \sum_{k=1}^{K} w_k \E{q_{\theta_k}(z_0)}{w_{\theta}(z_0) \nabla_{\theta_k} \log q_{\theta}(z_0)}.
	\end{split}
\end{equation}
We can find an alternative expression for the latter term. Starting with the latter term in \Cref{eq2:mixture_grad_term2_thetak}, we first use the definition in \Cref{eq2:mixture_pdf} and then apply the log-derivative trick, yielding the following expression,
\begin{equation}
	\begin{split}
		& \sum_{k=1}^{K} w_k \E{q_{\theta_k}(z_0)}{w_{\theta}(z_0) \nabla_{\theta_k} \log q_{\theta}(z_0)}\\
		& = \E{q_{\theta}(z_0)}{w_{\theta}(z_0)\nabla_{\theta_k} \log q_{\theta}(z_0)} \\
		& = \E{q_{\theta}(z_0)}{w_{\theta}(z_0)\frac{1}{q_{\theta}(z_0)}w_k q_{\theta_k}(z_0) \nabla_{\theta_k} \log q_{\theta_k}(z_0)} \\
		& = w_k\E{q_{\theta_k}(z_0)}{w_{\theta}(z_0) \nabla_{\theta_k} \log q_{\theta_k}(z_0)}.
	\end{split}
\end{equation}

Finally, we obtain the gradient $\nabla_{w_k} \E{q_{\theta}^{(t)}(z)}{f_{\theta}(z)}$, taken w.r.t.\ the mixture weights. We apply the expression derived in the main paper,
\begin{equation}\label{eq2:mixture_grad_term2_wk}
	\begin{split}
		& \nabla_{w_k} \E{q_{\theta}^{(t)}(z)}{f_{\theta}(z)} = \E{q_{\theta}^{(t)}(z)}{-\nabla_{w_k}\log q_{\theta}(z)} \\
		& + \E{q_{\theta}(z_0)}{w_{\theta}(z_0) \nabla_{w_k} \log q_{\theta}(z_0)}.
	\end{split}
\end{equation}
We rewrite the first term in \Cref{eq2:mixture_grad_term2_wk} by taking the exact expectation with respect to the mixture indicator,
\begin{equation}
	\begin{split}
		& \E{q_{\theta}^{(t)}(z)}{-\nabla_{w_k}\log q_{\theta}(z)} \\
		& = \sum_{k^\prime=1}^{K} w_{k^\prime}\E{q_{\theta_{k^\prime}}(z_0)}{\E{Q^{(t)}(z\g z_0)}{-\frac{q_{\theta_k}(z)}{q_{\theta}(z)}}}.
	\end{split}
\end{equation}
We rewrite the second term in \Cref{eq2:mixture_grad_term2_wk} using the log-derivative trick,
\begin{equation}
	\begin{split}
		& \E{q_{\theta}(z_0)}{w_{\theta}(z_0) \nabla_{w_k} \log q_{\theta}(z_0)} \\
		& = \E{q_{\theta}(z_0)}{ w_{\theta}(z_0)\frac{1}{q_{\theta}(z_0)}q_{\theta_k}(z_0)\nabla_{w_k} \log q_{\theta_k}(z_0)} \\
		& = \E{q_{\theta_{k}}(z_0)}{ w_{\theta}(z_0)\nabla_{w_k} \log q_{\theta_k}(z_0)}.
	\end{split}
\end{equation}





\end{document}